\documentclass[sigconf]{acmart}

\usepackage{arydshln} % must place in the top to use \cdashline
\usepackage{amsmath}
\usepackage{amsfonts}
\usepackage{bm}
\usepackage{multirow} 
\usepackage{graphicx}
\usepackage{subcaption}
\usepackage{multirow}
\usepackage{color}
\usepackage{xcolor}
\usepackage{booktabs}
\usepackage{colortbl}
\usepackage{bbding}
\usepackage{enumitem}

\AtBeginDocument{%
  }
\usepackage{booktabs}
\usepackage{multirow}
\usepackage{bm}
\usepackage{makecell} %%%%
\usepackage{colortbl}
\def\mycolor{\cellcolor[HTML]{CFEFFF}}

\usepackage{xspace}

\newcommand{\modelname}{SCING\xspace}
\usepackage{subcaption}

\begin{document}

\title{SCING:Towards More Efficient and Robust Person Re-Identification through Selective Cross-modal Prompt Tuning}

\author{Yunfei Xie$^{\dag}$}
\orcid{0009-0000-8853-1751}
\affiliation{%
  \institution{Huazhong University of Science and Technology}
  \city{Wuhan}
  \state{Hubei}
  \country{China}
  }
  \email{xieyunfei01@gmail.com}
\author{Yuxuan Cheng$^{\dag}$}
\orcid{0009-0007-2568-1905}
\affiliation{%
  \institution{Huazhong Agricultural University}
  \city{Wuhan}
  \state{Hubei}
  \country{China}
  }
  \email{hanxuanwxss@gmail.com}
\author{Juncheng Wu}
\orcid{xxxx}
\affiliation{%
  \institution{University of California, Santa Cruz}
  \city{Santa Cruz}
  \state{California}
  \country{U.S.A.}
  }
  \email{jwu418@ucsc.edu}
\author{Haoyu Zhang}
\orcid{0009-0006-0459-9892}
\affiliation{%
  \institution{City University of Hong Kong (Dongguan)}
  \city{Dongguan}
  \state{Guangdong}
  \country{China}
  }
  \email{haoyu.zhang@cityu-dg.edu.cn}
\author{Yuyin Zhou}
\orcid{0000-0003-2232-9563}
\affiliation{%
  \institution{University of California, Santa Cruz}
  \city{Santa Cruz}
  \state{California}
  \country{U.S.A.}
  }
  \email{zhouyuyiner@gmail.com}
\author{Shoudong Han}
\orcid{0000-0003-0572-4748}
\authornote{Corresponding author}
\affiliation{%
  \institution{Huazhong University of Science and Technology}
  \city{Wuhan}
  \state{Hubei}
  \country{China}
  }
  \email{shoudonghan@hust.edu.cn}

\renewcommand{\shortauthors}{Yunfei Xie et al.}



%  \includegraphics[width=\textwidth]%{sampleteaser}
  %\caption{Seattle Mariners at Spring Training, 2010.}
  %\Description{Enjoying the baseball game from the third-base
  %seats. Ichiro Suzuki preparing to bat.}
  %\label{fig:teaser}


\begin{abstract}
Recent advancements in adapting vision-language pre-training models like CLIP for person re-identification (ReID) tasks often rely on complex adapter design or modality-specific tuning while neglecting cross-modal interaction, leading to high computational costs or suboptimal alignment. To address these limitations, we propose a simple yet effective framework named \textbf{S}elective \textbf{C}ross-modal Prompt Tun\textbf{ing}(\textbf{SCING}) that enhances cross-modal alignment and robustness against real-world perturbations. Our method introduces two key innovations: Firstly, we proposed Selective Visual Prompt Fusion (SVIP), a lightweight module that dynamically injects discriminative visual features into text prompts via a cross-modal gating mechanism. Moreover, the proposed Perturbation-Driven Consistency Alignment (PDCA) is a dual-path training strategy that enforces invariant feature alignment under random image perturbations by regularizing consistency between original and augmented cross-modal embeddings. Extensive experiments
are conducted on several popular benchmarks covering Market1501, DukeMTMC-ReID, Occluded-Duke, Occluded-REID, and P-DukeMTMC, which demonstrate the impressive performance of the proposed method. Notably, our framework eliminates heavy adapters while maintaining efficient inference, achieving an optimal trade-off between performance and computational overhead. The code will be released upon acceptance.
\end{abstract}

\maketitle

\section{Introduction} 

\begin{figure}
    \centering
    \begin{subfigure}{0.15\textwidth}
        \centering
        \includegraphics[width=\linewidth]{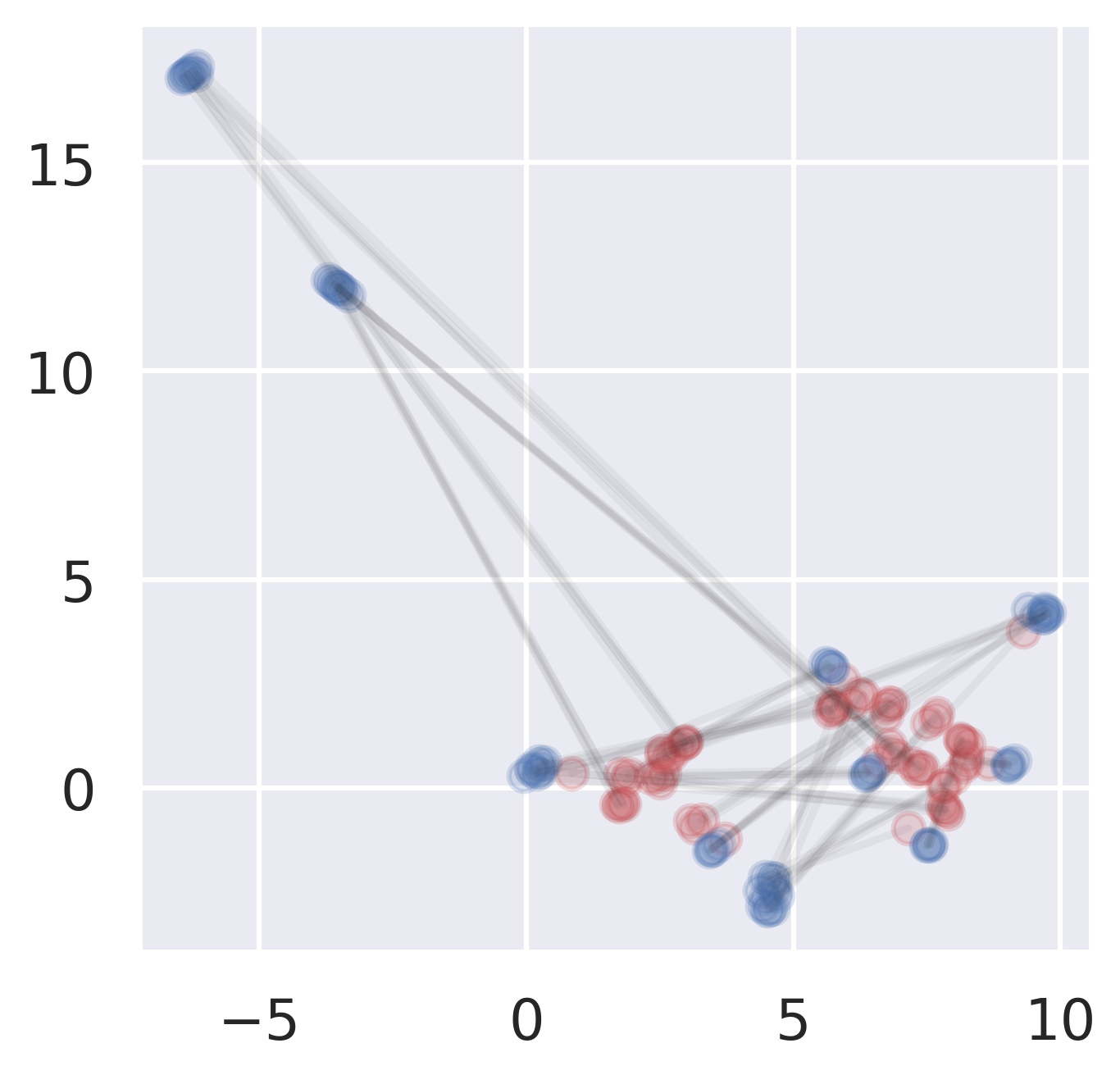}
        \caption{CLIP-ReID}
        \label{fig:teaser_a}
    \end{subfigure}
    \begin{subfigure}{0.15\textwidth}
        \centering
        \includegraphics[width=\linewidth]{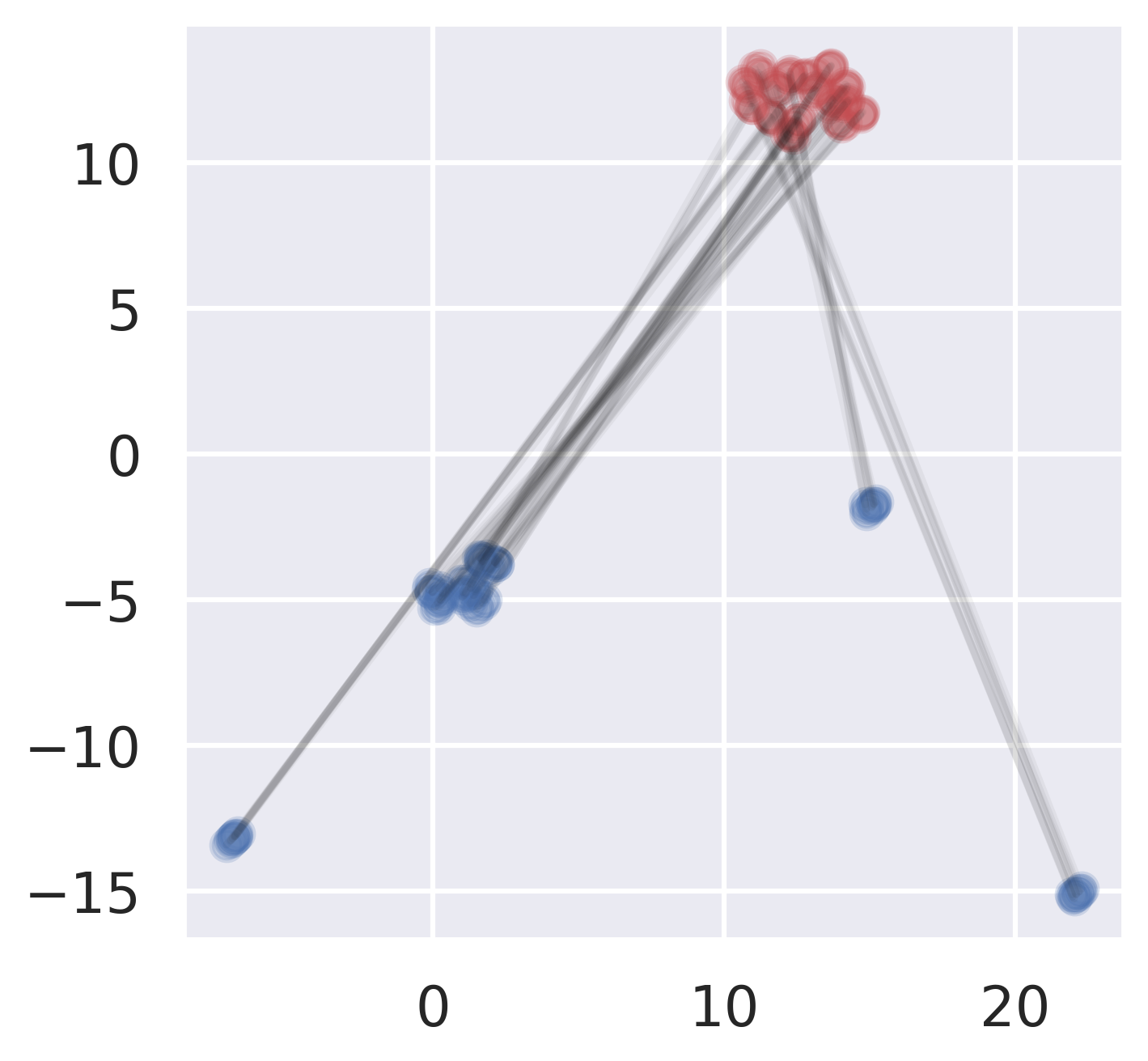}
        \caption{CoCoOp}
        \label{fig:teaser_b}
    \end{subfigure}
    \begin{subfigure}{0.15\textwidth}
        \centering
        \includegraphics[width=\linewidth]{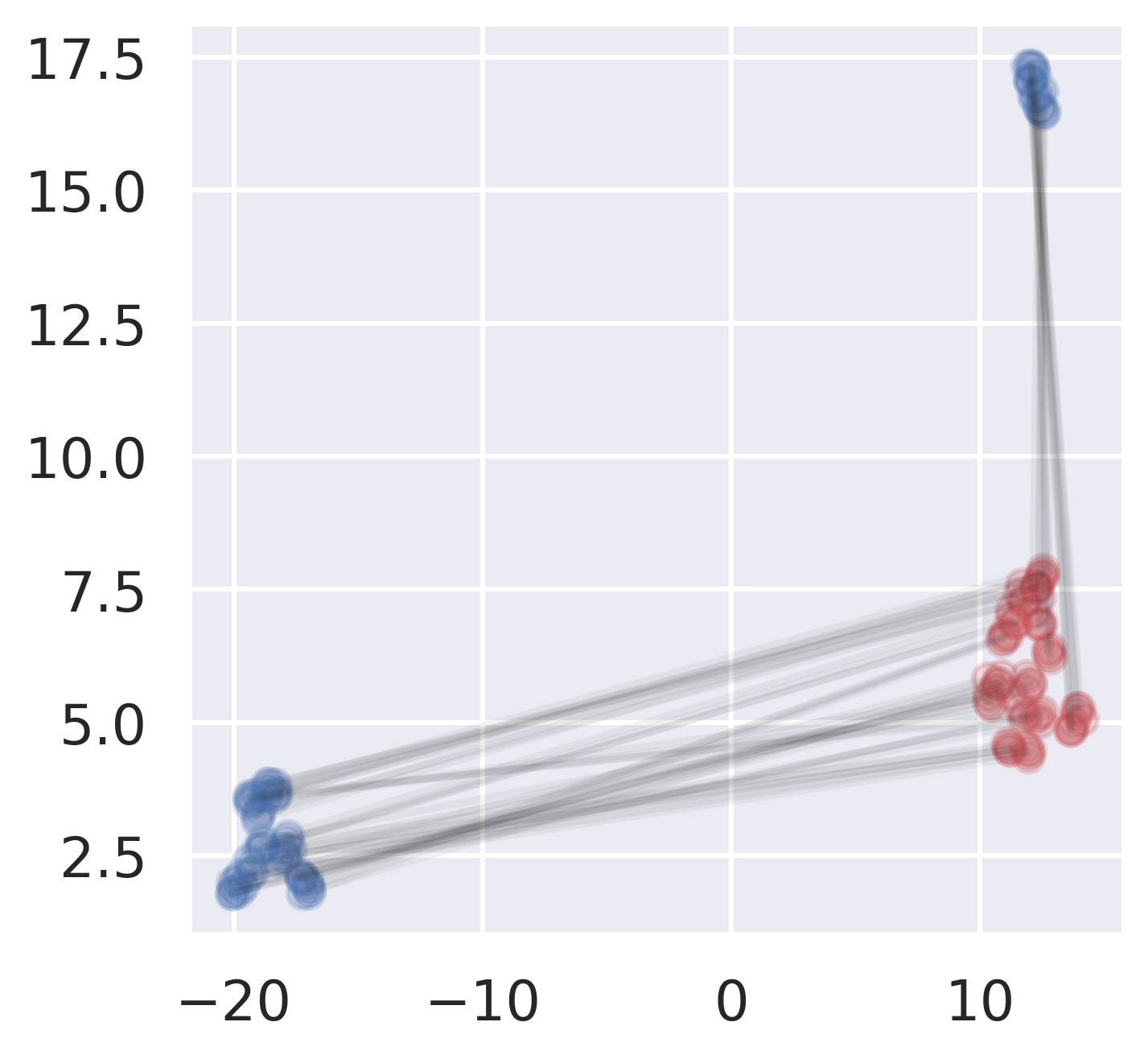}
        \caption{Ours}
        \label{fig:teaser_c}
    \end{subfigure}    
    \caption{UMAP visualization~\cite{sainburg2021parametric} shows the image-text modality gap using red dots for image embeddings and blue dots for paired text embeddings connected by grey lines. We comparing modality gap among CLIP-ReID, CoCoOp, and our approach. CLIP-ReID (a) and CoCoOp (b) exhibit a significant modality gap between person images and their corresponding textual descriptions, while our model (c) significantly reduces this gap, achieving improved image-text alignment.}
    \label{fig:teaser}
\end{figure}

\begin{figure}
    \centering
    \includegraphics[width=1\linewidth]{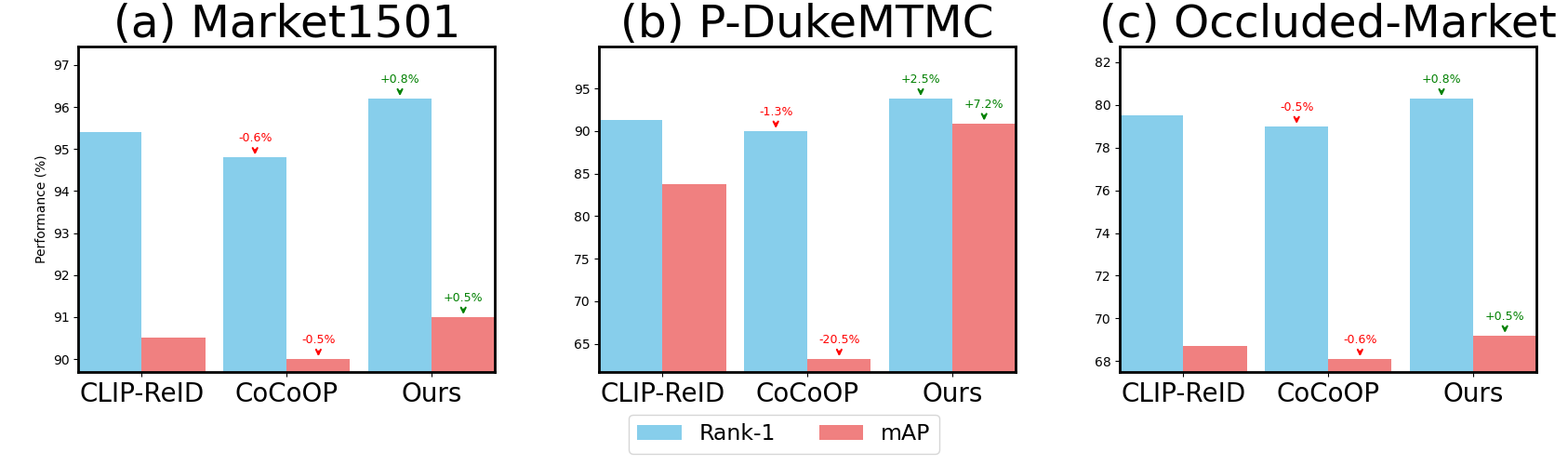}
    \caption{Performance comparison of different methods among CLIP-ReID, CoCoOp, and Ours in three different datasets. The graphics present that the CoCoOp method cannot be directly transferred to ReID, while the proposed Selective Cross-modal Prompt Tuning (\modelname) can effectively improve the performance of CLIP-based methods on ReID tasks.}
    \label{fig:bar}
\end{figure}
\label{sec:intro}
Person Re-Identification (ReID) aims to retrieve images of the same individual across non-overlapping cameras, a significant yet valuable capability for intelligent surveillance and security systems. Unlike standard image retrieval and other related tasks(e.g., instance retrieval~\cite{chen2022deep}, fine-grained classification~\cite{wei2021fine}), ReID faces unique challenges in unconstrained real-world scenarios~\cite{peng2023deep}: severe occlusions (e.g., partial body coverage by objects or crowds), cross-view appearance discrepancies (e.g., lighting variations, pose changes), and cluttered backgrounds that obscure identity-critical features.

Recent years, many research works have turned to get more robust and discriminable representation through utilizing the strong power from pre-trained multi-modal foundation models, leading to better downstream performance in ReID tasks. 
CLIP-ReID~\cite{li2023clip} was the initial approach to use pre-trained CLIP with prompt learning methods for ReID tasks. However, this method lacks cross-modal interaction, which results in suboptimal alignment between image and text features (Fig.~\ref{fig:teaser}).
To solve this problem, Conditional Context Optimization (CoCoOp)~\cite{Zhou2022CoCoop} approaches image-text interaction by compressing images into single visual tokens and combining them equally with text prompts (Fig.~\ref{fig:fuse_comparison} (a)). Although this works well for natural image classification, ReID scenarios frequently contain occluded objects, complex backgrounds, and viewpoint changes. This causes the simple fusion strategy in CoCoOp to introduce identity-irrelevant noise and leads to reduced performance in ReID tasks, as shown in Fig.~\ref{fig:bar}. ProFD~\cite{cui2024profd} investigated manually-designed local prompt-guided feature disentangling by implementing a complex adaptor and part-specific prompt design and using a pre-trained segmentation model (.~\ref{fig:fuse_comparison} (b)). This approach relies heavily on complex modality-specific adapter modules and external labels, resulting in higher computational costs and inflexible manual processes, as demonstrated in Table~\ref{tab:efficiency}.

To better bridge the cross-modal gaps in the ReID field, we propose a novel ReID adaptation framework named \textbf{S}elective \textbf{C}ross-modal Prompt Tun\textbf{ing} (\textbf{\modelname}) that establishes targeted cross-modal interaction and robust perturbation alignment. First, the proposed Selective Visual Prompt Fusion (SVIP) dynamically integrates discriminative local visual cues into part of text learnable tokens via a simple weighted gating mechanism, filtering out background noise while preserving identity-critical semantics. Second, the Perturbation-Driven Consistency Alignment (PDCA) maintains cross-modal consistency between the learnable text prompts fused with perturbed samples and the original image representation, enhancing modal interaction while improving the robustness of the model to real-world perturbations.

Extensive experiments on Market1501 \cite{Masson2019ASO} and DukeMTMC-ReID \cite{Zhang2017MultiTargetMT}, as well as occluded datasets, namely Occluded-Duke \cite{miao2019pose}, Occluded-ReID \cite{Zhuo2018OccludedPR}, P-DukeMTMC \cite{Zhang2017MultiTargetMT} and Occluded-Market~\cite{zhang2024focus} demonstrate surpassing performance. In particular, our method outperforms the CLIP-ReID approaches on several popular benchmarks with negligible parameter increase at the inference stage.

Overall, the contributions of this paper lie in the following aspects:
\begin{itemize}
    \item We propose a simple yet efficient ReID framework named SCING, combining two key components: Selective Visual Prompt Fusion (SVIP) and Perturbation-Driven Consistency Alignment (PDCA) to address modal sub-optimal alignment due to the lack of modal interaction and the perturbation in real-world scenarios.
    \item The framework is lightweight with negligible parameter increase during inference compared to the vision backbone, making it practical for practical deployment.  
    \item Our method achieves superior performance on many popular benchmarks in both holistic ReID and occluded ReID tasks.
\end{itemize}
\section{Related Work} 
%\subsection{Person Re-identification}
%Person Re-identification task aims to match a person's images from different non-overlapped cameras. 
\subsection{Vision-Language Learning}
Vision-language models have revolutionized various computer vision tasks by establishing cross-modal alignment between visual and textual modalities. Pre-trained models like CLIP~\cite{radford2021learning} learn transferable representations by training dual encoders on massive image-text pairs, projecting both into a shared embedding space. While these models demonstrate impressive zero-shot capabilities~\cite{kirillov2023segment,yang2021learning}, adapting them to specialized tasks poses significant challenges.

Recent approaches have addressed these limitations through innovative adaptation strategies. Prompt-based methods~\cite{Zhou2021Coop,huang2024learning} employ learnable tokens to create task-specific textual representations, effectively transferring generalization capabilities to downstream domains. Similarly, feature disentangling techniques~\cite{cheng2024disentangled} introduce specialized prompts to guide representation learning in challenging scenarios, combining spatial and semantic attention mechanisms to generate well-aligned features despite missing visual information. Other methods like lightweight adapters~\cite{gao2024clip} utilize compact modules to transfer pre-trained knowledge to downstream tasks with minimal parameter updates. These approaches further incorporate knowledge preservation techniques~\cite{li2024encapsulating}, such as self-distillation with memory banks~\cite{wang2024prompt}, to maintain the rich pre-trained knowledge while adapting to downstream tasks.

\subsection{ReID With Pre-trained Vision-Language Models}
In recent years, with the sharp development of large-scale pre-trained models~\cite{radford2021learning,li2022blip,wang2024qwen2,liu2023visual}, much research~\cite{li2023clip,cui2024profd,zuo2024ufinebench,lu2025clip, zhang2024magic} in ReID communities has turned its interest to using the strong generalization power from vision-language models to solve key points like crowd occlusion, perturbation from the true world, and so on. ~\cite{zhai2024multi} proposed a simple yet efficient two-stage training strategy by distributing learnable text prompt tokens for each class's person images with the CLIP model first. Similarly, ProFD\cite{cui2024profd} introduces part-specific prompts to guide feature disentangling in occluded scenarios, combining spatial and semantic attention mechanisms to represent well-aligned part features. CLIP3DReID~\cite{liu2024distilling} leverages CLIP’s knowledge distillation to align language-guided 3D shape priors with visual cues, enabling multi-level feature alignment between local attributes and global identity representations. MP-ReID~\cite{zhai2024multi}designed a multi-prompt process and achieved excellent performance on ReID tasks by using LLM to generate a variety of text prompts combined with learnable tokens.

As different from the former works,NAM ~\cite{tan2024harnessing} transfer their focus to the Multimodal Large Language Model (MLLM) and design a text-to-image ReID framework utilizing the image perception ability of MLLM. 

Overall, despite extensive efforts in vision-language model-based ReID, existing approaches often overlook cross-modal interaction mechanisms during learnable prompt or adapter tuning, relying instead on isolated modality-specific adaptations or overly complex adapter architectures.

\section{Methodology}

\begin{figure*}[t]
  \centering
  \includegraphics[page=1, width=0.9\textwidth]{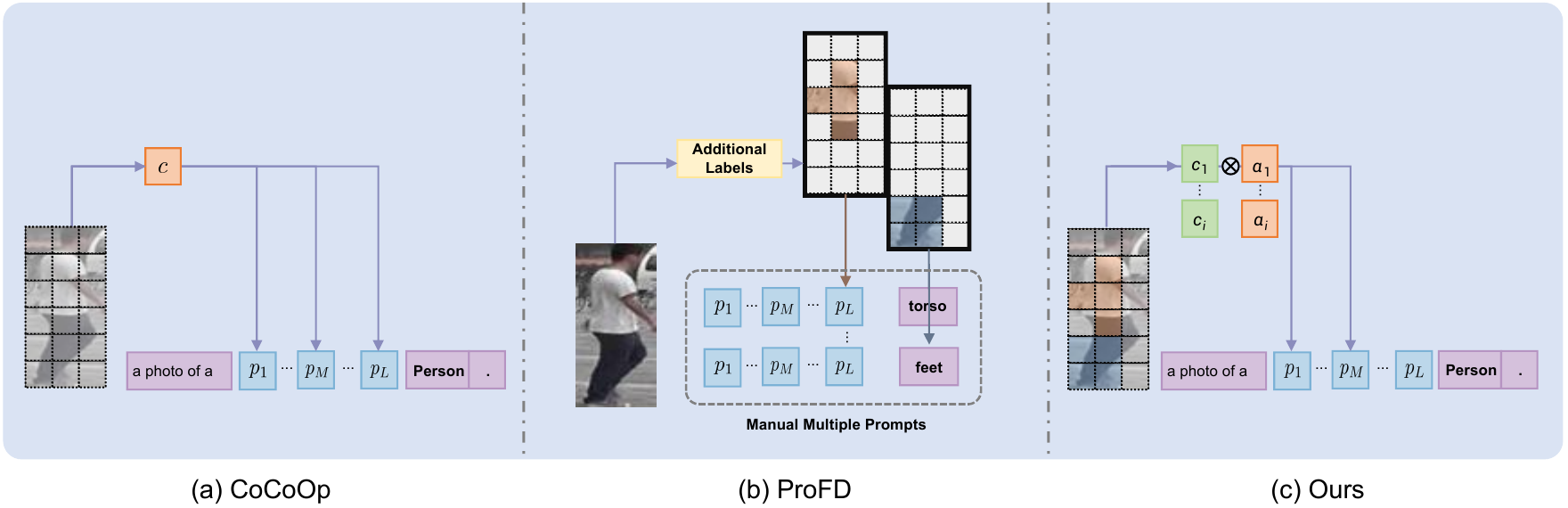}
\caption{
\textbf{Three approaches for integrating visual features into text prompts.} 
CoCoOp (a) compresses image information into a single visual token and fuses it equally with text prompts; ProFD (b) uses additional mask labels and manually designed part-specific prompts to align visual and text features; \modelname (Ours) (c) proposes a Selective Visual Prompt Fusion (SVIP) module that dynamically fuses relevant visual information to text tokens without requiring additional masks or prompts.}

  \label{fig:fuse_comparison}
\end{figure*}

\begin{figure*}[t]
  \centering
  \includegraphics[page=1, width=0.9\textwidth]{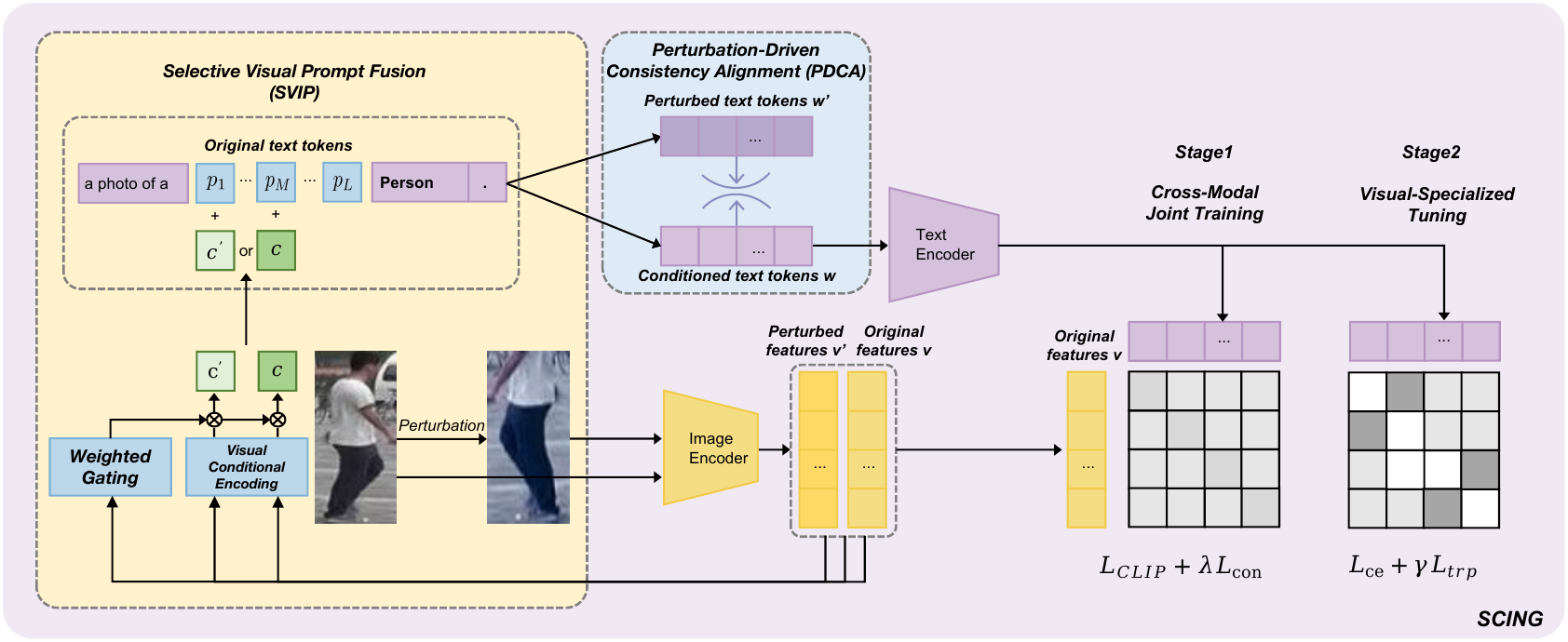}
\caption{
\textbf{Model framework.} For a given image and its learnable text description, we first use Selective Visual Prompt Fusion (SVIP) to enable cross-modal interaction. Next, we generate perturbed samples and apply Perturbation-Driven Consistency Alignment (PDCA) to improve the model's robustness against real-world perturbations. 
Our training includes two stages: first, we optimize learnable visual fusion text tokens for each class, and second, we further tune the visual encoder for better optimization.
}
  \label{fig:framework}
\end{figure*}
\subsection{A Review of CLIP and CoCoOp}
\textbf{Contrastive Language-Image Pre-training} (CLIP)~\cite{radford2021learning} pioneers a dual-stream architecture that learns semantically aligned representations between images and text. It consists of two key components:
\begin{itemize}
\item \textbf{Image Encoder}: A vision backbone (ResNet~\cite{he2016deep} or ViT~\cite{dosovitskiy2020image}) mapping image $I$ to feature vector $\bm{x} \in \mathbb{R}^d$.
\item \textbf{Text Encoder}: A Transformer~\cite{vaswani2017attention} network converting text prompts into embeddings $\{\bm{w}_i\}_{i=1}^K \in \mathbb{R}^d$.
\end{itemize}

\noindent During pre-training, CLIP optimizes a \textit{bidirectional contrastive loss} to align matched image-text pairs in a shared embedding space. For a batch of $N$ pairs $\{(I_i, T_i)\}_{i=1}^N$, the loss is:

\begin{equation}
\mathcal{L}_{\text{CLIP}} = -\frac{1}{2N} \sum_{i=1}^N \left[ \log \frac{e^{\langle \bm{x}_i, \bm{w}_i \rangle / \tau}}{\sum_{j=1}^N e^{\langle \bm{x}_i, \bm{w}_j \rangle / \tau}} + \log \frac{e^{\langle \bm{x}_i, \bm{w}_i \rangle / \tau}}{\sum_{j=1}^N e^{\langle \bm{x}_j, \bm{w}_i \rangle / \tau}} \right],
\end{equation}

\noindent where $\tau$ is a learnable temperature parameter, and $\langle\cdot,\cdot\rangle$ denotes cosine similarity function. 

For zero-shot inference, CLIP generates text embeddings $\{\bm{w}_i\}_{i=1}^K$ by encoding template-based prompts (e.g., \textit{``a photo of a \{class\}''}) with $K$ class names. The classification probability for image $I$ is computed as:

\begin{equation}
p(y|I) = \frac{\exp\left(\langle\bm{x}, \bm{w}_y\rangle/\tau\right)}{\sum_{i=1}^K \exp\left(\langle\bm{x}, \bm{w}_i\rangle/\tau\right)},
\label{eq:clip_zero_shot}
\end{equation}
~\textbf{Conditional Context Optimization}(CoCoOp)~\cite{Zhou2022CoCoop} extends CLIP by introducing \textit{image-conditioned prompts} to address the rigidity of hand-crafted templates. Unlike CoOp~\cite{Zhou2021Coop}, which learns static prompt vectors $\{\bm{p}_i\}_{i=1}^L$ for all images, CoCoOp generates dynamic prompts conditioned on each input image $I$:
\begin{equation}
    c = \textit{MLP}(\bm{x}_g)
\end{equation}
\begin{equation}
\bm{p}_i(I) = \bm{\mu}_i + c
\end{equation}

\noindent where $\bm{\mu}_i$ are learnable basis vectors, $\bm{x}_g$ is the global image feature from CLIP's visual encoder, and $\textit{MLP}(\cdot)$ is a two-layer perceptron with an activation function. The final text embedding $\bm{w}_k$ for class $k$ is then:

\begin{equation}
\bm{w}_k = F_{\text{t}}\left(\left[\bm{p}_1(I), \bm{p}_2(I), \dots, \bm{p}_L(I), \text{``class: } C_k \text{''}\right]\right).
\end{equation}
CoCoOp~\cite{Zhou2022CoCoop} partially bridges the modality gap between global visual patterns and text prompts by conditioning prompts on image features. This dynamic adaptation allows text embeddings $\bm{w_k}$ to encode instance-specific semantics (e.g., scene context or object attributes) and enhances alignment consistency compared to CLIP’s fixed prompts and CoOp’s~\cite{Zhou2021Coop} single-modal optimization, which tunes static text-side context.

\subsection{Framework Overviews}
Existing CLIP-based ReID models typically rely on complex adapter designs and modality-specific tuning, which independently optimize visual or textual encoders while neglecting cross-modal interaction. To address this limitation, we propose a simple yet effective framework, SCING, as shown in Fig.~\ref{fig:framework}. Our framework comprises two key components:
\begin{itemize}
    \item \textbf{S}elective \textbf{Vi}sual \textbf{P}rompt Fusion (\textbf{SVIP}): A lightweight module that selectively injects discriminative visual features into learnable text prompts via a cross-modal gating mechanism. Unlike CoCoOp’s indiscriminate fusion of global image characteristics with learnable tokens on the text side, our method strategically aggregates critical local characteristics (e.g., faces, gestures) while filtering out identity-irrelevant noise (e.g., background clutter).
    \item \textbf{P}erturbation-\textbf{D}riven \textbf{C}onsistency \textbf{A}lignment(\textbf{PDCA}): A dual-path training strategy that enforces invariant feature alignment under random perturbations. By minimizing the similarity between fusion prompts generated from perturbed and original images, the model learns to focus on identity-critical regions and drops real-world distortions.
\end{itemize}
%Compared to prior arts, our framework eliminates the need for heavy modality-specific adapters while achieving superior cross-modal alignment, as validated in Sec.~\ref{？？}. 

\subsection{Selective Visual Prompt Fusion}
\label{sec:svip}

To bridge the modality gap in prompt learning and enhance cross-modal interaction, we propose a \textbf{Selective Visual Prompt Fusion (SVIP)} module that dynamically fuses critical visual clues with learnable text tokens, as illustrated in Fig.~\ref{fig:fuse_comparison}.

\subsubsection{Learnable Prompt Initialization}
We initialize $ L $ randomly sampled tokens in the text prompt:
\begin{equation}
    \bm{P} = [\bm{p}_1, \bm{p}_2, \dots, \bm{p}_L] \in \mathbb{R}^{L \times d}
\end{equation}

where the first $ M $ tokens ($ M < L $) participate in visual-textual interaction. 
\subsubsection{Visual Condition Encoding}
Given input image $ I $, the CLIP visual encoder $ F_v $ extracts image features:

\begin{equation}
    \bm{V} = F_v(I) \in \mathbb{R}^{D}
\end{equation}

A compact visual condition is generated via:

\begin{equation}
\bm{c} = \text{MLP}(V) \in \mathbb{R}^d
\label{eq:cond}
\end{equation}

where $ \text{MLP} $ denotes a two-layer perceptron with ReLU activation.

\subsubsection{Feature Selection Mechanism}
We design a weighted gating module to select discriminative visual features:

\begin{equation}
\bm{\alpha} = \sigma\left(\bm{W}_s \cdot \bm{V} + \bm{b}_s\right) \in [0,1]^d
\label{eq:gate}
\end{equation}

where $ \sigma $ is the sigmoid function, $ \bm{W}_s \in \mathbb{R}^{D \times d} $ and $ \bm{b}_s \in \mathbb{R}^d $ are learnable parameters. The resulting $\bm{\alpha} = [\alpha_1, \alpha_2, \ldots, \alpha_m]$ contains $M$ individual weights, where each $\alpha_i \in [0,1]$ corresponds to the importance of the $i$-th visual condition.

\subsubsection{Cross-Modal Token Fusion}
The visual condition is adaptively fused into text prompts:

\begin{equation}
\bm{p}_i^{\text{svip}} = \bm{p}_i  + \bm{c_i} \odot \bm{\alpha_i}, \quad \forall i \in \{1, 2, \dots, M\}
\label{eq:fusion}
\end{equation}

The final text embedding for class $ k $ is computed as:

\begin{equation}
\bm{w}_k = F_t\left(\left[\text{``a photo of a }~\bm{p}_1^{\text{svip}}, \dots, \bm{p}_M^{\text{svip}}, \bm{p}_{M+1}, \dots, \bm{p}_L ~\text{person} \text{''}\right]\right)
\label{eq:text_emb}
\end{equation}
where $w_k$ represents the text feature, $F_t$ denotes the text encoder from CLIP.

\subsection{Perturbation-Driven Consistency Alignment}
\label{sub:consistency}

To enhance robustness against real-world perturbations, we propose a \textbf{Perturbation-Driven Consistency Loss} that aligns augmented fused text embeddings with original visual features. As shown in Fig.~\ref{fig:framework}, the method operates as follows:

\subsubsection{Perturbation Generation}
Given an input image $I$, we generate two augmented views through stochastic transformations:
\begin{equation}
I' = \mathcal{T}(I), \quad I'' = \mathcal{T}(I)
\end{equation}
where $\mathcal{T}(\cdot)$ denotes the random perturbation function. The detail of random perturbation function is in Sec. \ref{sec:imple}
\subsubsection{Feature Extraction and Modality Fusion}
For each perturbed image $I^{(m)} \in \{I', I''\}$ we first use image encoder and selective visual prompt fusion method to generate corresponding perturbed text embedding $\bm{w} \in \{w', w''\}$.

\subsubsection{Consistency Loss Formulation}
To ensure consistent text representations despite image perturbations, we compute the cosine similarity between text embeddings derived from the original and perturbed images. This encourages the text encoder to generate semantically consistent representations for the same person under various real-world occlusion and view transformation.

The cosine similarity between two embeddings is defined as:
\begin{equation}
\text{cos}(\bm{w}_i, \bm{w}_j) = \frac{\bm{w}_i^\top \bm{w}_j}{\|\bm{w}_i\|\|\bm{w}_j\|}
\end{equation}

The final consistency loss aggregates the similarity measures between all pairs of text embeddings:
\begin{equation}
\mathcal{L}_{\text{con}} = 1 - \frac{1}{3} \left[ \text{cos}(\bm{w}, \bm{w}') + \text{cos}(\bm{w}, \bm{w}'') + \text{cos}(\bm{w}', \bm{w}'') \right]
\label{eq:pdca_final}
\end{equation}

This consistency loss serves a critical purpose: it ensures that text embeddings remain semantically consistent when derived from the same identity under different perturbations. By maximizing the similarity between text representations, our model becomes robust to real-world variations like partial occlusions, pose changes, and viewpoint shifts.

\subsection{Training and Inference}
\label{sub:training}

Our framework adopts a two-stage training paradigm to balance cross-modal alignment and task-specific discriminability.

\subsubsection{Stage-1: Cross-Modal Joint Training}
In the first stage, we jointly optimize all parameters in both visual and textual streams with two loss components:

CLIP Loss: Inherited from Eq.~(1), maintains basic image-text alignment:
\begin{equation}
\mathcal{L}_{\text{CLIP}} = \mathcal{L}_{\text{t2i}} + \mathcal{L}_{\text{i2t}}
\end{equation}

The total objective integrates both components with balancing factor $\lambda$:
\begin{equation}
\mathcal{L}_{\text{stage1}} = \mathcal{L}_{\text{clip}} + \lambda \mathcal{L}_{\text{con}}
\label{eq:stage1_loss}
\end{equation}
where $\mathcal{L}_{\text{con}}$ denotes the proposed consistency loss from Eq.~\ref{eq:pdca_final}
%\subsubsection{Stage-2: Visual-Specific Fine-tuning}
%In the second stage, we freeze textual components and focus on visual discriminability. In particular, we use fixed text features from different classes without visual prompt fusion,  and fine-tune all parameters of the visual encoder through simple cross-entropy loss:
%\begin{equation}
%    L_{ce}=F
%\end{equation}
%where $I_k$ denotes input image and $w_k$ represents matching

\subsubsection{Stage 2: Visual-Specialized Tuning}
In the second stage, we transition to unimodal visual optimization by freezing all textual components (\textit{text encoder $F_v$
 , SVIP prompts, fusion parameters}), and only use fixed text prompts $w_i$ from different classes. While we conduct full-parameter tuning of the visual encoder $F_v$
to maximize identity-specific discriminability without complex adaptor design, given an input image $I_{k}$ with identity label $y=k$, we compute:

\begin{equation}
\mathcal{L}_{\text{ce}} = -\log\frac{\exp(\langle F_v(I_k), \bm{w}k\rangle)}{\sum_{i=1}^K \exp(\langle F_v(I_k), \bm{w}_i\rangle)}
\label{eq:final_ce}
\end{equation}
where $K$ denotes the number of classes in datasets.

At the same time, a simple cross-modal triplet loss is adopted to further compact vision feature representation as the below equation:
\begin{equation}
\mathcal{L}_{\text{trp}} = \max\big( \langle F_v(I_k), \bm{w}_k \rangle - \langle F_v(I_k), F_v(I_n) \rangle + \alpha, 0 \big)
\end{equation}
Here, $F_v(I_n)$ represents the hardest negative visual feature in the batch, selected via:

\begin{equation}
    I_n = \arg\min_{\substack{I_j \in \mathcal{B} \\ y_j \neq k}} \langle F_v(I_k), F_v(I_j) \rangle
\end{equation}
Where $\mathcal{B}$ is the current batch and $\alpha=0.2$ is the margin. 
Overall, the loss function for Stage 2 is as follows:
\begin{equation}
    \mathcal{L}_{\text{stage2}} = \mathcal{L}_{\text{ce}} + \gamma\mathcal{L}_{\text{trp}}
\end{equation}
And $\gamma$ denotes the balance weights between each loss functions.

\subsubsection{Inference}

During Inference, our method only uses the vision backbone from CLIP without any other parameters and using $\bm{g}_p$ as the global descriptor for each image and performs the Person Re-Identification task. We adopt cosine distance as the metric to measure the similarity between the query descriptor $\mathbf{g}^{\text{r}}$ and each target descriptor $\mathbf{g}^{\text{t}}$:
\begin{equation}
\Phi(\mathbf{g}^{\text{r}}, \mathbf{g}^{\text{t}}) = 1 - \frac{\mathbf{g}^{\text{r}} \cdot \mathbf{g}^{\text{t}}}{\|\mathbf{g}^{\text{r}}\| \|\mathbf{g}^{\text{t}}\|}
\end{equation}

\section{Experiment}
\subsection{Datasets and Metrics}
\textbf{Datasets.}
To highlight that our model maintains performance on holistic datasets and demonstrates improvement on occluded datasets, we selected the following datasets:  holistic datasets, including Market1501 \cite{Masson2019ASO} and DukeMTMC-ReID \cite{Zhang2017MultiTargetMT}, as well as occluded datasets, namely Occluded-Duke \cite{miao2019pose}, Occluded-ReID \cite{Zhuo2018OccludedPR}, P-DukeMTMC \cite{Zhang2017MultiTargetMT} and Occluded-Market~\cite{zhang2024focus}. The details are shown as follows:

\begin{itemize}
    \item \textbf{Market1501:} Comprising 32,668 labeled images of 1,501 identities captured by 6 cameras, this dataset is divided into a training set with 12,936 images representing 751 identities, used exclusively for model pre-training.
    \item \textbf{DukeMTMC-ReID:} This dataset consists of 36,411 images showcasing 1,404 identities from 8 camera. It includes 16,522 training images, 17,661 gallery images, and 2,228 queries.
    \item \textbf{Occluded-Duke:} Containing 15,618 training images, 2,210 occluded query images, and 17,661 gallery images, this dataset is a subset of DukeMTMC-ReID, featuring occluded images and excluding some overlapping ones.
    \item \textbf{Occluded-ReID:} Captured by mobile camera equipment on campus, this dataset includes 2,000 annotated images belonging to 200 identities. Each person in the dataset is represented by 5 full-body images and 5 occluded images with various types of occlusions.
    \item \textbf{P-DukeMTMC:} Derived from the DukeMTMC-ReID dataset, this modified version comprises 12,927 images (665 identities) in the training set, 2,163 images (634 identities) for querying, and 9,053 images in the gallery set.
    
     \item \textbf{Occluded-Market:} Formed by combining and re-partitioning MARS~\cite{zheng2016mars} and Market-1501~\cite{Masson2019ASO}. Its training set of it contains 9287 images with 780 IDs, the query set contains 2343 images with 533 IDs, and the gallery set contains 15913 images with 751 IDs. Same as Occluded-DukeMTMC, it also follows the setting that all the images in the query set are occluded images, but the proportion of occluded images in the training set is 63\%, which is much higher than that in Occluded-DukeMTMC.    
\end{itemize}

\noindent \textbf{Evaluation Metrics.}
Following established conventions in the ReID community, we assess performance using two standard metrics: the Cumulative Matching Characteristics (CMC) at Rank-1 and the Mean Average Precision (mAP). Evaluations are conducted without employing re-ranking \cite{zhong2017re} in a single-query setting.

\subsection{Implementation Details}
\label{sec:imple}
%The training procedure follows the approach in CLIP-ReID~\cite{li2023clip} and ProFD~\cite{cui2024profd}. 

Consistent with CLIP-ReID~\cite{li2023clip}, we adopt a two-stage training process. In the first stage, only the learnable text tokens $\rm[X]_1[X]_2...[X]_M$ are optimized, which combine with visual conditions through a feature selection mechanism. In the second stage, we fix these learned text tokens and optimize only the visual encoder.

For both training and inference, input images are resized to $256 \times 128$ with a patch size of $16 \times 16$. During training, we apply data augmentation to person images, including random flipping, random erasing, and random cropping, each with a 50\% probability. The batch size for both training stages is set to 64, with 4 images per person. We use the Adam optimizer with a weight decay of 0.0005. The learning rate begins at 5e-5 and decreases by a factor of 0.1 at epochs 30 and 50. The model is trained for 120 epochs for each training stage.

Our perturbation strategy combines both image and feature-level techniques. Image-level perturbations include random flipping, erasing, cropping, and occlusion. For feature-level perturbation, we apply dropout with 50\% probability to the feature maps generated by the visual backbone.

\subsection{Baseline}
We comprehensively evaluated representative methods of both non-CLIP-based models and CLIP-based models in six different datasets covering the holistic Person ReID task and the more challenging occluded Person ReID task. Specifically, for the holistic person ReID task, we compared methods including: MGN~\cite{wang2018learning}, PCB~\cite{sun2018beyond}, PCB+RPP~\cite{sun2018beyond}, VPM~\cite{sun2019perceive}, Circle~\cite{sun2020circle}, ISP~\cite{zhu2020identity},
TransReID~\cite{he2021transreid},
DC-Former*~\cite{li2023dc},
PGFA~\cite{miao2019pose},
PGFL-KD~\cite{zheng2021pose},
HOReID~\cite{wang2020high},
MHSA~\cite{tan2022mhsa},
BPBreID~\cite{somers2023body},
RGANet~\cite{he2023region},
PAT~\cite{li2021diverse},
FED~\cite{wang2022feature},
DPM~\cite{tan2022dynamic},
FRT~\cite{xu2022learning},
PFD~\cite{wang2022pose},
SAP~\cite{jia2023semi},
CLIP-ReID~\cite{li2023clip},
CoCoOp~\cite{zhou2022conditional}, and
ProFD~\cite{cui2024profd}.

For the occluded Person ReID task, we evaluated approaches such as Part-Aligned~\cite{zhao2017deeply}, PCB~\cite{sun2018beyond}, Adver Occluded~\cite{huang2018adversarially}, PVPM~\cite{gao2020pose}, PGFA~\cite{miao2019pose}, HOReID~\cite{wang2020high}, GASM~\cite{he2020guided}, VAN~\cite{yang2021learning}, OAMN~\cite{chen2021occlude}, PGFL-KD~\cite{zheng2021pose}, PAT~\cite{li2021diverse}, DRL-Net~\cite{jia2022learning}, TransReID~\cite{he2021transreid}, BPBreID~\cite{somers2023body}, MHSA~\cite{tan2022mhsa}, FED~\cite{wang2022feature}, MSDPA~\cite{cheng2022more}, FRT~\cite{xu2022learning}, SAP~\cite{jia2023semi}, DPM~\cite{tan2022dynamic}, RGANet~\cite{he2023region}, CLIP-ReID~\cite{li2023clip}, CoCoOp~\cite{zhou2022conditional}, ProFD~\cite{cui2024profd}.
%\subsection{Comparison with the State-of-the-Art}
\subsection{Evaluation on Holistic Person ReID Dataset}
As shown in Table~\ref{tab:2}, we conducted comprehensive comparative experiments on the Market1501 and DukeMTMC-ReID datasets.
\noindent \subsubsection{Evaluation on Market1501}
As shown in Table~\ref{tab:2}, our method achieves state-of-the-art performance on the Market1501 dataset, surpassing existing CLIP-based and non-CLIP-based approaches in Rank-1 accuracy and mAP. Specifically, compared to the baseline CLIP-ReID model, our method demonstrates significant improvements of +0.8\% in Rank-1 (96.2\% vs. 95.4\%) and +0.5\% in mAP (91.0\% vs. 90.5\%), highlighting the effectiveness of enhancing cross-modal interaction during fine-tuning. Notably, while methods like CoCoOp attempt to integrate global visual features with learnable text tokens, their indiscriminate fusion strategy—as discussed in Section~\ref{sec:intro}—risks overfitting case-level background noise in ReID datasets, which compromises the compact feature representations critical for retrieval tasks. Furthermore, our approach outperforms ProFD~\cite{cui2024profd}, which relies on pre-trained segmentation models and handcrafted local prompts, suggesting that our lightweight and streamlined design achieves competitive performance without requiring complex architectural modifications. This underscores the potential of selective cross-modal prompt tuning as a more straightforward yet powerful paradigm for ReID.

\noindent \subsubsection{Evaluation on DukeMTMC-ReID}

On DukeMTMC-ReID, our approach attains competitive results, securing second place in both metrics and closely following the top-performing method. Compared to the baseline CLIP-ReID, the proposed method achieves comprehensive leadership, improving Rank-1 accuracy by 0.5 percentage points and mAP by 0.6 percentage points (reaching 91.3\% and 83.7\%, respectively). Notably, applying CoCoOp in ReID, which indiscriminately fuses global image features and learnable text prompts, actually leads to a performance decrease. Specifically, its mAP of 82.7\% on DukeMTMC-ReID is lower than the CLIP-ReID baseline (83.1\%) in the mAP metric. These experiments demonstrate the superior performance of our method and provide evidence for the effectiveness of the proposed selective visual prompt fusion.
\renewcommand{\multirowsetup}{\centering}
\begin{table}[t]
  \begin{center}
   \caption{\label{tab:2}Performance comparison of the holistic ReID problem on the Market1501 and DukeMTMC-ReID. Methods are categorized into Non CLIP-based and CLIP-based approaches. $*$ indicates the backbone is with an overlapping stride setting, stride size $s_o=12$.}
\setlength{\tabcolsep}{1.8mm}
\renewcommand{\arraystretch}{1.0}{
\scalebox{0.77}{
  \begin{tabular}{ l|l|cc|cc}
\Xhline{1.0pt}
& &\multicolumn{2}{c|}{Market1501} &  
\multicolumn{2}{c}{DukeMTMC-ReID}\\
\multirow{-2}{*}{\parbox{1.2cm}{\centering Back-\\bone}} & \multirow{-2}{*}{Method}& Rank-1 & mAP & Rank-1 & mAP\\
\hline\hline
\multirow{22}{*}{\parbox{1.2cm}{\centering Non\\CLIP-\\based}} 
& MGN~\cite{wang2018learning} 	& 95.7 & 86.9 & 88.7 & 78.4 \\
& PCB~\cite{sun2018beyond} 	& 92.3 & 77.4 & 81.7 & 66.1  \\
& PCB+RPP~\cite{sun2018beyond} 	& 93.8 &81.6 &83.3 &69.2  \\
& VPM~\cite{sun2019perceive} 	& 93.0 & 80.8 & 83.6 & 72.6 \\
& Circle~\cite{sun2020circle} 	& 94.2 & 84.9 & - & -  \\
& ISP~\cite{zhu2020identity} 	& 95.3 & 88.6 & 89.6 & 80.0  \\
& TransReID~\cite{he2021transreid}  & 95.2 & 88.9 & 90.7 &82.6\\
& DC-Former*~\cite{li2023dc}  & 96.0 & 90.4 & - & -\\
& PGFA~\cite{miao2019pose}	&91.2 &76.8 &82.6 &65.5 \\
& PGFL-KD~\cite{zheng2021pose} &95.3 &87.2 &89.6 &79.5  \\
& HOReID~\cite{wang2020high} 	  &94.2 &84.9 &86.9 &75.6  \\
& MHSA~\cite{tan2022mhsa} &94.6 &84.0 &87.3 &73.1\\
& BPBreID~\cite{somers2023body}  &95.1 &87.0 &89.6 &78.3\\
& RGANet~\cite{he2023region}  &95.5 & 89.8 & - & - \\
& PAT~\cite{li2021diverse} 	& 94.2 & 84.9 &  88.8 & 78.2\\
& FED~\cite{wang2022feature}  &95.0 &86.3 &89.4 &78.0 \\
& DPM*~\cite{tan2022dynamic} 	 & 95.5 & 89.7 & 91.0 & 82.6\\ 
& FRT~\cite{xu2022learning} 	 & 95.5 &  88.1 & 90.5 & 81.7 \\
& PFD*~\cite{wang2022pose}  & 95.5 & 89.7 & 91.2 & 83.2\\
& SAP* ~\cite{jia2023semi}  & \underline{96.0} & 90.5 & - & -\\
\hline
\multirow{4}{*}{\parbox{1.2cm}{\centering CLIP-\\based}} 
& CLIP-ReID~\cite{li2023clip}  &95.4 &90.5 & 90.8 &83.1 \\
& CoCoOp~\cite{zhou2022conditional}  &94.8 &90.0 & 90.8 &82.7 \\
& ProFD~~\cite{cui2024profd}	&95.6 &\underline{90.8}  & \textbf{92.1} & \textbf{84.0} \\
&\mycolor{\textbf{\modelname(Ours)}} &  \mycolor{\textbf{96.2}} & \mycolor{\textbf{91.0}} & \mycolor{\underline{91.3}} & \mycolor{\underline{83.7}} \\
\hline
  \end{tabular}}}
  \end{center}
\end{table}

\subsection{Evaluation on Occluded Person ReID Dataset.}

\begin{table}[t]
  \begin{center}
   \caption{\label{tab:1}Performance comparison of the occluded ReID problem on the Occluded-Duke, Occluded-ReID, P-DukeMTMC and Occluded-Market. Methods are categorized into Non CLIP-based and CLIP-based approaches.}
\setlength{\tabcolsep}{1.3mm}
\renewcommand{\arraystretch}{0.95}{
\scalebox{0.6}{
  \begin{tabular}{ l|l|c|c|c|c}
& & \multicolumn{1}{c|}{Occluded-Duke} & \multicolumn{1}{c|}{Occluded-ReID} & \multicolumn{1}{c|}{P-DukeMTMC} & \multicolumn{1}{c}{Occluded-Market}\\
\multirow{-2}{*}{\parbox{1.2cm}{\centering Back-\\bone}} & \multirow{-2}{*}{Method}& Rank-1 / mAP & Rank-1 / mAP & Rank-1 / mAP & Rank-1 / mAP\\
\hline\hline
\multirow{21}{*}{\parbox{1.2cm}{\centering Non\\CLIP-\\based}} 
& Part-Aligned~\cite{zhao2017deeply} & 28.8 / 20.2 & - / - & - / - & - / - \\
& PCB~\cite{sun2018beyond} & 42.6 / 33.7 & 41.3 / 38.9 & - / - & 66.0 / 49.4 \\
& Adver Occluded~\cite{huang2018adversarially} & 44.5 / 32.2 & - / - & - / - & - / - \\
& PVPM~\cite{gao2020pose} & 47.0 / 37.7 & 70.4 / 61.2 & 51.5/ 29.2 & 66.8 / 49.4 \\
& PGFA~\cite{miao2019pose} & 51.4 / 37.3 & - / - & 44.2 / 23.1 & 64.1 / 45.5 \\
& HOReID~\cite{wang2020high} & 55.1 / 43.8 & 80.3 / 70.2 & - / - & 64.9 / 49.3 \\
& GASM~\cite{he2020guided} & - / - & 74.5 / 65.6 & - / - & - / - \\
& VAN~\cite{yang2021learning} & 62.2 / 46.3 & - / - & - / - & - / - \\
& OAMN~\cite{chen2021occlude} & 62.6 / 46.1 & - / - & - / - & - / - \\
& PGFL-KD~\cite{zheng2021pose} & 63.0 / 54.1 & 80.7 / 70.3 & 81.1 / 64.2 & - / - \\
& PAT~\cite{li2021diverse} & 64.5 / 53.6 & 81.6 / 72.1 & - / - & - / - \\
& DRL-Net~\cite{jia2022learning} & 65.8 / 53.9 & - / - & - / - & - / - \\
& TransReID~\cite{he2021transreid} & 66.4 / 59.2 & - / - & - / - & 78.2 / 64.7 \\
& BPBreID~\cite{somers2023body} & 66.7 / 54.1 & 76.9 / 68.6 & 91.0 / 77.8 & - / - \\
& MHSA~\cite{tan2022mhsa} & 59.7 / 44.8 & - / - & 70.7 / 41.1 & - / - \\
& FED~\cite{wang2022feature} & 68.1 / 56.4 & 86.3 / 79.3 & - / - & 66.7 / 53.3 \\
& MSDPA~\cite{cheng2022more} & 70.4 / 61.7 & 81.9 / 77.5 & - / - & - / - \\
& FRT~\cite{xu2022learning} & 70.7 / 61.3 & 80.4 / 71.0 & - / - & - / - \\
& SAP* ~\cite{jia2023semi} & 70.0 / 62.2 & 83.0 / 76.8 & - /- & - / - \\
& DPM*~\cite{tan2022dynamic} & \underline{71.4} / 61.8 & 85.5 / 79.7 & - / - & - / - \\
& RGANet~\cite{he2023region} & \textbf{71.6} / 62.4 & 86.4 / 80.0 & - / - & - / - \\
\hline
\multirow{4}{*}{\parbox{1.2cm}{\centering CLIP-\\based}} 
& CLIP-ReID~\cite{li2023clip} & 67.2 / 60.3 & - / - & 91.3 / 83.7 & 79.5 / 68.7 \\ 
& CoCoOp~\cite{zhou2022conditional} & 70.0 / 62.4 & - /  - &  
90.0 / 83.2 & 79.0 / 68.1   \\ 
& ProFD~\cite{cui2024profd} & 70.6 / \underline{63.1} & \underline{92.3} / \underline{90.3} & \underline{92.8} / \textbf{84.7} & - / - \\
& \mycolor{\textbf{\modelname(Ours)}} & \mycolor{71.1 / \textbf{63.4}}& \mycolor{\textbf{93.8} / \textbf{90.9}} & \mycolor{\textbf{93.7} / \underline{84.4}} & \mycolor{\textbf{80.3} / \textbf{69.2}} \\
\hline
  \end{tabular}}}
  \end{center}
\end{table} 

%To further evident the efficiency of our method, we conduct more experiences on occluded ReID datasets. table~\ref{tab:1} demonstrates that
Occluded person re-identification (Occluded ReID) presents a more challenging scenario, where the goal is to retrieve individuals under severe occlusion, viewpoint variations, or partial observations. To rigorously evaluate the robustness of our method, we conduct comprehensive experiments on four Occluded ReID benchmarks: Occluded-Duke, Occluded-ReID, P-DukeMTMC, and Occluded-Market.

As shown in Table~\ref{tab:1}, our approach achieves surpassing performance among CLIP-based methods across all datasets while demonstrating competitive advantages over non-CLIP-based approaches in most scenarios.

\noindent\textbf{On Occluded-Duke dataset}, our method attains 71.1\% Rank-1 and 63.4\% mAP, surpassing the best CLIP-based competitor ProFD~\cite{cui2024profd} (70.6\% Rank-1 / 63.1\% mAP) by +0.5\% and +0.3\%, respectively. While the non-CLIP method RGANet ~\cite{he2023region} achieves a slightly higher Rank-1 (71.6\%), our method significantly outperforms it in mAP (+1.0\% over RGANet’s 62.4\%), highlighting the advantage of the proposed Perturbation-Driven Consistency Alignment loss based on selective cross-modal prompt tuning in improving retrieval consistency under occlusion.

\noindent\textbf{On Occluded-ReID and Occluded-Market datasets}, our method achieves unambiguous state-of-the-art performance across all existing methods. For Occluded-ReID, we attain 93.8\% Rank-1 and 90.9\% mAP, surpassing the strongest CLIP-based competitor ProFD~\cite{cui2024profd} (92.3\% / 90.3\%) by +1.5\% in Rank-1 and +0.6\% in mAP, while outperforming the best non-CLIP method FED~\cite{wang2022feature} (86.3\% / 79.3\%) by remarkable margins of +7.5\% and +11.6\%. Similarly, on Occluded-Market, our method achieves 80.3\% Rank-1 and 69.2\% mAP, exceeding both CLIP-ReID~\cite{li2023clip} (79.5\% / 68.7\%) and the top non-CLIP approach TransReID~\cite{he2021transreid} (78.2\% / 64.7\%) in both metrics. These results validate that our perturbation consistency paradigm effectively addresses extreme occlusion patterns without relying on auxiliary modules like pre-trained segmentation networks (ProFD) or the design of complex local feature perceptron (TransReID), establishing a new benchmark for occlusion-robust ReID.

\noindent\textbf{On P-DukeMTMC dataset}, our method demonstrates a balanced yet rank-prioritized performance: it achieves 93.7\% Rank-1, surpassing ProFD’s 92.8\% (+0.9\%), while its mAP (84.4\%) slightly trails ProFD’s 84.7\% (-0.3\%).  This trade-off suggests our design emphasizes rank-sensitive discriminability—critical for real-world retrieval systems—by avoiding ProFD’s segmentation-dependent local prompts, which may overfit to dataset-specific part annotations. Despite the marginal gap in mAP, our framework maintains competitive overall performance through selective cross-modal interaction, further highlighting its practical advantages in simplicity.

%\noindent \subsubsection{Evaluation on Occluded-Duke}
%\noindent \subsubsection{Evaluation on Occluded-ReID}
%\noindent \subsubsection{Evaluation on P-DukeMtMC}
%\noindent \subsubsection{Evaluation on Occluded-Market}
\subsection{Ablation Study}
\begin{table}[!t]
\centering
\caption{The Ablation Studies on Occluded-Duke dataset}
\scalebox{0.75}{
\begin{tabular}{cccccc}
\toprule
CLIP-ReID & metanet & \makecell[c]{Selective Visual \\ Prompt Fusion} & \makecell[c]{Perturbation-Driven \\ Consistency Alignment} & mAP & rank-1 \\
\midrule
$\checkmark$ & & & & 59.1 & 65.7 \\
$\checkmark$ & $\checkmark$ & & & 58.0 & 67.4 \\
$\checkmark$ &  & $\checkmark$ & & 62.1 & 69.6 \\
\mycolor{$\checkmark$} & \mycolor{ } & \mycolor{$\checkmark$} & \mycolor{$\checkmark$} & \mycolor{\textbf{63.4}} & \mycolor{\textbf{71.1}} \\
\bottomrule
\end{tabular}}
\label{tab:ablation_component}
\end{table}
To meticulously evaluate the individual contributions of our proposed components, namely the Selective Visual Prompt Fusion (SVIP) and the Perturbation-Driven Consistency Alignment (PDCA), we conduct a series of ablation experiments. As shown in table~\ref{tab:ablation_component}, systematically demonstrate the efficacy of each module within our framework using mAP and Rank-1 accuracy as evaluation metrics on the Occluded-Duke dataset.

We begin with our baseline configuration, denoted as CLIP-ReID, which represents a standard CLIP model prompt-tuned for the ReID task, achieving 59.1\% mAP and 65.7\% Rank-1 accuracy.

Next, we introduce the meta-net from CoCoOp~\cite{Zhou2022CoCoop} to enhance modal interaction during the tuning process. As discussed in Section~\ref{sec:intro}, indiscriminately fusing global image characteristics with text prompts can easily lead the model to overfit to instance-specific background noise, which is detrimental to learning the compact, identity-discriminative features required for the ReID task. The experimental results support this concern: while incorporating the meta-net slightly improves Rank-1 accuracy to 67.4\% (+1.7\% compared to the baseline), the mAP decreases to 58.0\% (-1.1\%). This mixed outcome suggests that directly applying CoCoOp's indiscriminate fusion strategy, while facilitating some cross-modal interaction, may indeed capture identity-irrelevant noise, hindering overall precision in the context of ReID.

Subsequently, we integrate the core contribution of our Selective Visual Prompt Fusion (SVIP) module (Section~\ref{sec:svip}) (Row 3). This involves adding the feature selection mechanism (Eq.~\ref{eq:gate}) and the adaptive cross-modal token fusion (Eq.~\ref{eq:fusion}) to the visually conditioned prompts. The results show a significant improvement over the indiscriminate fusion approach, boosting the mAP to 62.1\% (+4.1\% compared to the model which only introduces the meta-net) and Rank-1 accuracy to 69.6\% (+2.2\% compared to Row 2). This substantial gain underscores the importance of SVIP's ability to selectively integrate discriminative visual cues into the text prompts while filtering out irrelevant information, effectively bridging the modality gap and enhancing cross-modal interaction in a more targeted manner suitable for ReID.

Finally, we introduce the Perturbation-Driven Consistency Alignment (PDCA) strategy (Section~\ref{sub:consistency}) into stage 1 alongside SVIP (Row 4), representing our full proposed model. By enforcing consistency among the text embeddings generated via selective fusion from the perturbed image and original 
image features using $L_{con}$ (Eq.~\ref{eq:pdca_final}), the model's robustness is further enhanced. This leads to the best performance, achieving 63.4\% mAP (+4.3\% over baseline) and 71.1\% Rank-1 (+5.4\% over baseline). This final increment validates the effectiveness of PDCA in encouraging the model to learn identity-invariant representations that are robust to common real-world variations like occlusion,

\subsection{Efficiency Comparison with Other CLIP-based Model}
\begin{table}[t]
  \begin{center}
    \caption{Efficiency analysis of different CLIP-based ReID methods. Avg. Rank-1 and Avg. mAP are computed across Occluded-Duke, Occluded-ReID, and P-DukeMTMC datasets.}
    \label{tab:efficiency}
    \setlength{\tabcolsep}{1.2mm}
    \renewcommand{\arraystretch}{1.0}
    \scalebox{0.9}{
    \begin{tabular}{l|c|c|c|c}
      \hline
      Method & Parameters (M) & FLOPs (G) & Avg. Rank-1 & Avg. mAP \\
      \hline\hline
      CLIP-ReID & 126.55 & 24.14 & 67.20 & 60.30 \\      
      CoCoOp & 126.59 & 24.14 & - & - \\
      ProFD & 138.47 & 38.11 & 85.23 & 79.37 \\
      \mycolor{\textbf{\modelname~(Ours)}} & \mycolor{126.95} & \mycolor{24.14} & \mycolor{\textbf{86.20}} & \mycolor{\textbf{79.57}} \\      
      \hline
    \end{tabular}}
  \end{center}
\end{table}
We evaluate the efficiency of SCING against other CLIP-based methods in Table~\ref{tab:efficiency}, focusing on parameters (M), FLOPs (G), and average Rank-1/mAP accuracy in Occluded-Duke, Occluded-ReID, and P-DukeMTMC datasets.

To be detailed, compared to CLIP-ReID and CoCoOp, our method (SCING) shows a negligible increase in parameters (126.95\, M vs. \textasciitilde126.6\,M) while maintaining identical FLOPs (24.14\,G). Crucially, this comes with a substantial performance boost, achieving 86.20\% Rank-1 and 79.57\% mAP, far exceeding CLIP-ReID's results (67.20\% / 60.30\%). At the same time, when compared with the high-performing ProFD, SCING demonstrates significant efficiency gains. It utilizes considerably fewer parameters (126.95\, M vs. 138.47\ ,M) and requires substantially fewer FLOPs (24.14\, G vs. 38.11\, G). Despite being much lighter, our method achieves slightly superior performance in both Rank-1 (86.20\% vs. 85.23\%) and mAP (79.57\% vs. 79.37\%).

In summary, SCING achieves the efficiency-performance balance, outperforming existing CLIP-based methods while maintaining or significantly reducing computational requirements.

\subsection{Visualization of \modelname}

\begin{figure}[t]
  \centering
  \includegraphics[page=1, width=0.4\textwidth]{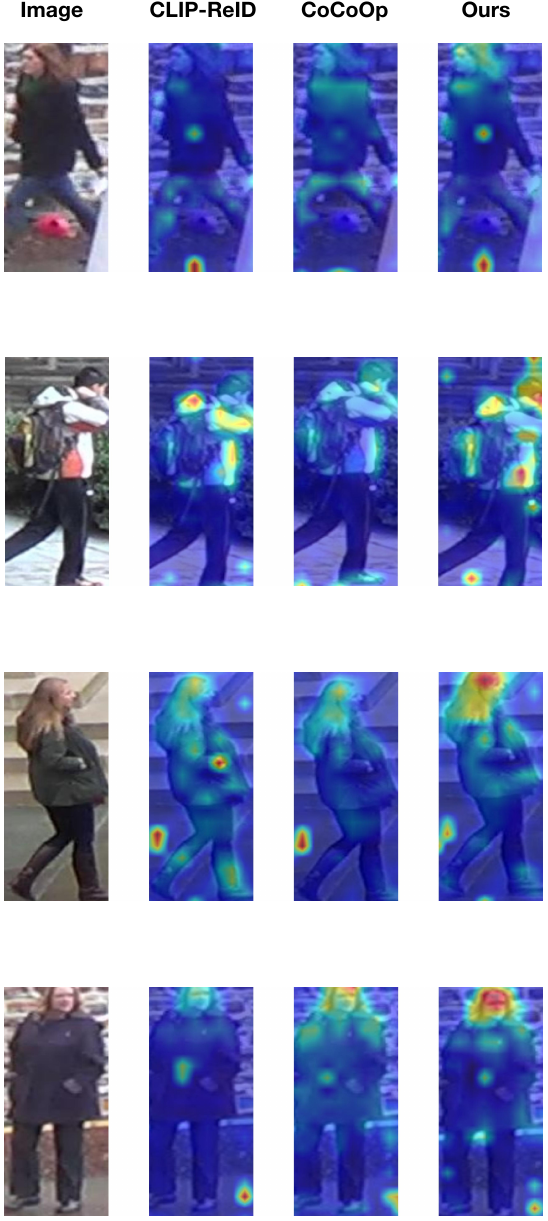}
\caption{
\textbf{Visualization of different prompt learning models on visual saliency maps.} Compared with other methods, \modelname focus on a more comprehensive area.}
  \label{fig:visual}
\end{figure}
As shown in Fig~\ref{fig:visual}, we perform visualization experiments using the gradcam method~\cite{selvaraju2020grad} to show the focused areas of the model. Both CLIP-ReID, CoCoOp and our \modelname focus on local areas, ignoring other details about the human body, while \modelname will focus on a more comprehensive area. For instance, in the first row, CLIP-ReID almost completely ignores identity-critical facial and hair features, focusing solely on clothing attributes - such bias could hinder practical person re-identification in uncontrolled environments. Similarly, the second row reveals CLIP-ReID's persistent deficiency in capturing key facial information, further validating the necessity of our proposed selective visual prompt fusion strategy. Notably, the third row from CoCoOp indicates that indiscriminate fusion of instance-level visual-textual prompts risks capturing irrelevant background elements while overlooking identity-related features, which fundamentally limits its effectiveness for ReID tasks. In contrast, our method effectively concentrates on identity-sensitive head details (e.g., facial features, hairstyles) while simultaneously capturing distinctive clothing textures, demonstrating strong potential for real-world person re-identification applications.
\section{Conclusion}
In this paper, we introduced SCING, a simple yet effective framework designed to enhance cross-modal interaction in CLIP-based Person Re-identification. SCING integrates two key components: a Selective Visual Prompt Fusion (SVIP) module and a Perturbation-Driven Consistency Alignment (PDCA) strategy. Extensive evaluations across six popular ReID benchmarks demonstrate the framework's versatility and efficacy, consistently achieving leading performance and highlighting its potential for robust real-world application

%\begin{acks}

%\end{acks}

%\section{Acknowledgments}
\bibliographystyle{ACM-Reference-Format}
\bibliography{camera_ready}

\appendix

\end{document}